\documentclass[11pt]{article}
\usepackage{acl}
\usepackage{times}
\usepackage{latexsym}
\usepackage{graphicx}
\usepackage{amsmath}
\usepackage{booktabs}
\usepackage[most]{tcolorbox}
\usepackage{enumitem}
\usepackage{multirow}
\usepackage[T1]{fontenc}
\usepackage[utf8]{inputenc}
\usepackage{microtype}
\usepackage{adjustbox}
\usepackage{caption}
\usepackage{titlesec}
\titlespacing*{\subsection}{0pt}{0.8ex plus 0.2ex minus 0.2ex}{0.3ex plus 0.1ex minus 0.1ex}
\usepackage{amssymb}

\usepackage[skip=0.3em]{caption} 

\newcommand{\Gngram}[2]{G_{#1}\!\left(#2\right)}    

\newcommand{\baseone}{\texttt{Direct-Prompting}}

\newcommand{\basetwo}{\texttt{Ground Every Sentence}}

\newcommand{\mone}{\texttt{LLaMA-3.1-8B }}

\newcommand{\mtwo}{\texttt{LLaMA-3.3-70B }}

\newcommand{\method}{\texttt{{EvidenT}}}

\title{\method: Building Trustworthy Enterprise Assistants through Evidence Groundedness and Traceability}

\author{
Anubha Kabra \And
Katie Jooyoung Kim \And
Colin Zhiwei Kou \And
Helene Sajer
\AND
Yimei Fan\hspace{29.73pt}Radomir Cisar\hspace{29.73pt}Heather Greenhalgh\hspace{29.73pt}Gabriel Martinez Vidiri
\AND
Bloomberg \\
\fontsize{9.7}{13.948}\selectfont
\texttt{\{akabra16, ckou9, yfan258, rcisar, hgreenhalgh1, gmartinezvi1\}@bloomberg.net} \\
\fontsize{9.7}{13.948}\selectfont
\texttt{jk4534@columbia.edu \quad helene.sajer.pro@gmail.com}
}

\begin{document}
\maketitle

\begin{abstract}
Enterprise AI assistants must produce responses that are verifiable and traceable to source evidence. However, retrieval-augmented generation (RAG) over heterogeneous enterprise data can suffer from citation drift, unsupported content, and weak source traceability. We present \texttt{\method} (\texttt{T} = Trust + Transparency + Traceability), a lightweight pipeline that verifies extracted evidence against retrieved documents before answer generation, without model retraining. EvidenT combines structured passage extraction with deterministic lexical alignment to filter unsupported content, correct citation drift, and preserve source-span traceability. On approximately 500 real enterprise queries, \method{} improves gold-source hit rate by an average of 29\% over prompting baselines, produces no citations to non-retrieved \texttt{urls}, and achieves near-saturated answer-to-source lexical coverage. 
\end{abstract}

\section{Introduction}

Recent advances in question answering (QA) with large language models (LLMs) have demonstrated impressive capabilities. Retrieval-augmented generation (RAG) pipelines, which pair LLMs with document retrievers to generate responses with inline citations, have achieved strong results on web-scale benchmarks \citep{novikova2017need,laban2022summac,kryscinski2020factcc}. However, applying these techniques to enterprise data presents additional challenges, particularly when source documents are heterogeneous, noisy, and inconsistently structured. In enterprise domains such as finance, healthcare, and law, factual precision and transparent source verification are particularly important \cite{chen2024survey}. Additionally, enterprise content is often fragmented across long, noisy, and inconsistently structured documents, making it harder for systems to locate and present the most relevant information accurately \cite{umar1999enterprise}. In our pilot deployment, we observed responses containing claims that were not well aligned with cited source content, as well as citations that were difficult to verify \citep{joren2025sufficient, choubey2025benchmarking, packowski2024optimizing}.

\begin{table*}[h]
\centering
\small
\resizebox{\textwidth}{!}{
\begin{tabular}{lccc}
\toprule
\textbf{Error Type} & \textbf{Description} & \textbf{Root Cause} & \textbf{Impact} \\
\midrule
Hallucinated Links & Nonexistent citations or \texttt{urls} & Pattern-based generation & Erodes trust \\
Citation Drift & Cited passage doesn’t support claim & Misaligned grounding & Reduces factual reliability \\
Limited Traceability & Hard to locate cited text & Buried content, weak anchors & Lowers transparency \\
\bottomrule
\end{tabular}
}
\caption{Key limitations of the pilot enterprise assistant deployment.}
\label{tab:limitations}
\end{table*}

To better understand these challenges, we conducted a month-long pilot deployment of an enterprise assistant built using a standard RAG architecture (see Appendix~\ref{app:pilot} for details). The assistant was deployed to 55 users spanning roles such as operations, legal, and support, and processed approximately 4,000 real-world queries. The resulting dataset was used to evaluate our proposed approach. Prompts used for this pilot are available in Appendix \ref{app:prompts}.

\subsection{Limitations of Pilot Enterprise Assistant Deployment}
\textbf{Groundedness} refers to whether generated claims are \textit{supported} by source documents. \textbf{Traceability} refers to the ability to associate generated content with \textit{localized supporting spans} in the source material. The pilot exposed three recurring citation and traceability failure modes in this setting (Table \ref{tab:limitations}).

\vspace{\itemsep}
\noindent\textbf{Invalid Citation Links:} The assistant generated plausible-looking \texttt{urls} that were not present in the retrieved enterprise sources.

\vspace{\itemsep}
\noindent\textbf{Citation Drift:} A valid document was cited, but the cited document did not contain the source text corresponding to the generated content.

\vspace{\itemsep}
\noindent\textbf{Limited Traceability:} Even when the cited document was appropriate, locating the relevant supporting text within long or unstructured documents could be difficult.

\vspace{\baselineskip}

\noindent These challenges led us to investigate the following research questions.

\textbf{RQ1:} Can we improve citation grounding and prevent invalid citation \texttt{urls} without model retraining?

\textbf{RQ2:} Can generated answers be reliably traced to localized source spans in noisy and heterogeneous enterprise documents?

We propose a lightweight pipeline designed to improve citation reliability and source traceability in extractive enterprise question answering. The approach operates without model retraining and can be applied on top of existing retrieval and generation components. The pipeline extracts candidate evidence, verifies it against retrieved source documents using lexical alignment, and generates answers from the resulting verified passages with associated source citations. This design prioritizes verifiability and source traceability under the operational constraints of enterprise deployment. On an evaluation set of approximately 500 queries derived from the pilot, \method{} improves gold-source hit rate by an average of 29\% over prompting baselines, produces no citations to non-retrieved \texttt{urls}, and achieves near-saturated lexical answer-to-source coverage.

\section{Previous Work}
\label{sec:limitations}

\subsection{Fine-tuning and Domain Adaptation of Generative Models}
Fine-tuning large language models has been extensively explored to enhance grounding and factual accuracy in retrieval augmented generation (RAG) pipelines \cite{huang2024learning, penzkofer2024evaluating, zhang2024raft}. However, fine-tuning requires large quantities of domain-specific labeled or synthetic data, which are often unavailable in enterprise contexts. Although fine-tuned models may yield more coherent responses, they still hallucinate or over-generalize under weak retrieval, as shown in \citep{lee2025finetune} and \citep{soudani2024fine}. Moreover, fine-tuning primarily improves fluency rather than factual traceability \cite{ghosal2024understanding}, leaving outputs difficult to map to supporting source spans. Prior work \citep{huang2024learning} shows that while model-level optimization can lower hallucination frequency, it fails to guarantee transparent citation alignment or user-verifiable provenance, limiting its utility in domains requiring citation-level accountability.

\begin{figure*}[t]
    \centering
    \includegraphics[width=0.8\textwidth]{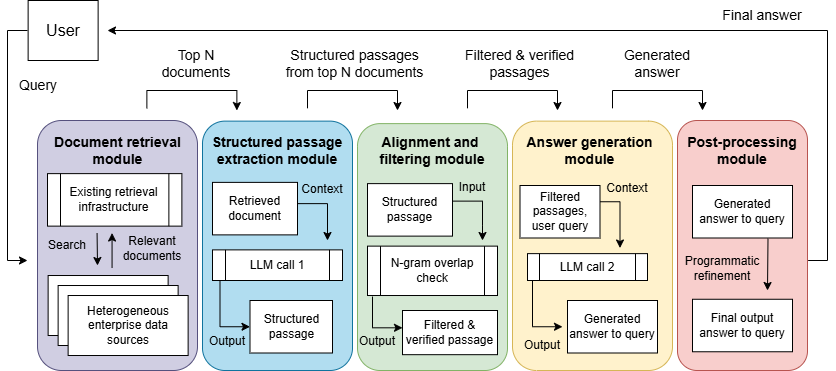}
    \caption{Complete modular pipeline.}
    \label{fig:pipeline}
\end{figure*}

\subsection{Basic RAG Systems}
Despite widespread use in grounded question answering, RAG systems continue to exhibit weaknesses in heterogeneous or noisy enterprise environments \cite{chen2024benchmarking}. The retrieval stage remains a major bottleneck: most retrievers assume clean, well-indexed corpora, whereas enterprise repositories are fragmented and inconsistent. Comprehensive reviews \citep{sharma2025retrieval, brown2025systematic} identify retrieval noise, redundancy, and incomplete indexing as leading sources of downstream hallucinations. Even when relevant materials are retrieved, \textit{citation drift} often occurs, where models cite correct documents but extract unsupported passages \citep{patel2024factuality, huang2024learning}. Because retrieval and generation are typically decoupled, existing pipelines lack explicit mechanisms to ensure that each claim is grounded in an identifiable source span. As a result, citation precision remains low, undermining user trust and auditability in enterprise deployments.

\subsection{Lack of Focus on Complete Traceability}
In RAG and LLM-based question answering, \textbf{complete traceability}, linking each generated factual statement to its exact supporting span, is essential for enterprise use. While prior work \cite{rashkin2021increasing, xu2025citeeval} introduces citation metrics focused on precision and recall, these do not capture the span-level grounding required for verifiable outputs. Furthermore, most evaluations \cite{gao2023enabling} rely on public benchmarks with short, homogeneous texts that differ markedly from the long and unstructured documents found in enterprise corpora. Consequently, current systems lack fine-grained provenance and remain limited in supporting transparent and verifiable generation.


\section{Our Approach}

Figure \ref{fig:pipeline} presents the pipeline components, and Figure \ref{fig:visual} illustrates the end-to-end query-processing workflow. The pipeline includes the following components.
\subsection{Document Retrieval Module}
This module operates via an automated federated retrieval process. Upon receiving a query, the system acts as a federated orchestrator, triggering parallel API calls to multiple pre-configured, independent data sources (e.g., wikis, policy manuals, and technical documentation). Because these enterprise repositories are managed independently, we treat each backend as a distinct ``black box''. The pipeline interfaces directly with each source's native retrieval infrastructure (whether lexical, dense, or hybrid) to aggregate the top \(N\) relevant documents into a unified context. This retriever-agnostic design ensures the system scales flexibly to new data sources in real time without requiring costly index re-alignment or manual intervention.

\subsection{Structured Passage Extraction Module}\label{subsec: extraction}
 This module extracts candidate passages from retrieved documents using an LLM with structured output formatting. The prompt is designed to return verbatim spans from the document context (see Appendix \ref{evident prompts}). Each passage is represented in the following JSON structure:

{\small
\begin{verbatim}
{
  "passage_id": "<passage_id>",
  "url": "<url>",
  "content": "<passage_content>"
}
\end{verbatim}
}

We take advantage of the fact that LLMs achieve near-perfect performance on verbatim span extraction from long-context source documents \citep{hsieh2024ruler} and structurally straightforward format transformations \citep{yang2025structeval}.  

\subsection{Alignment and  Filtering Module}\label{subsec: entailment}

 This module filters hallucinated passages based on n-gram overlap with their source documents. For each passage, we retrieve the full content of the source document using the provided \texttt{url} field. We then compute the n-gram overlap (typically \(n=5\)) between the passage  and the document content.
Let $p$ be an extracted passage and $d$ its cited document.
Define the $n$-gram overlap ratio as 
\begin{equation}
\small
\mathrm{overlap}_n(p,d) =
\frac{\lvert \Gngram{n}{p} \cap \Gngram{n}{d} \rvert}
     {\lvert \Gngram{n}{p} \rvert}.
\end{equation}
where $G_n(p)$ is the multiset of $n$-grams in $p$. We define the filtering action as follows:

\begin{itemize}[nosep,topsep=0pt,leftmargin=*] 
    \item If \textbf{overlap} \(>\) \textbf{threshold}: the passage is retained with the original URL.
    \item If \textbf{0} \(<\) \textbf{overlap} \(\leq\) \textbf{threshold}: the passage is truncated to retain only the overlapping portion; the original \texttt{url} is preserved.
    \item If \textbf{overlap} \(= 0\): we suspect citation drift. We iterate over all other retrieved documents to find one with overlap \(>\) threshold. If found, the passage \texttt{url} is replaced with the correct one.
    \item If none of these conditions apply, we drop the passage from the generated JSON.
\end{itemize}

We rely on lexical matching because our enterprise data is out of distribution with respect to the training data of most semantic models, because our design prioritizes verifiability over abstraction. Based on our experiments, the n-gram with a threshold of 0.7 performs best for our setting. This overlap threshold is configurable. In lower-risk or less latency-sensitive deployments, practitioners may tune the threshold or incorporate semantic or hybrid matching strategies. For the ablation on our data please see Appendix \ref{sec:hyperparameter_ablation}.

\begin{figure*}[t]
    \centering
    \includegraphics[width=\textwidth]{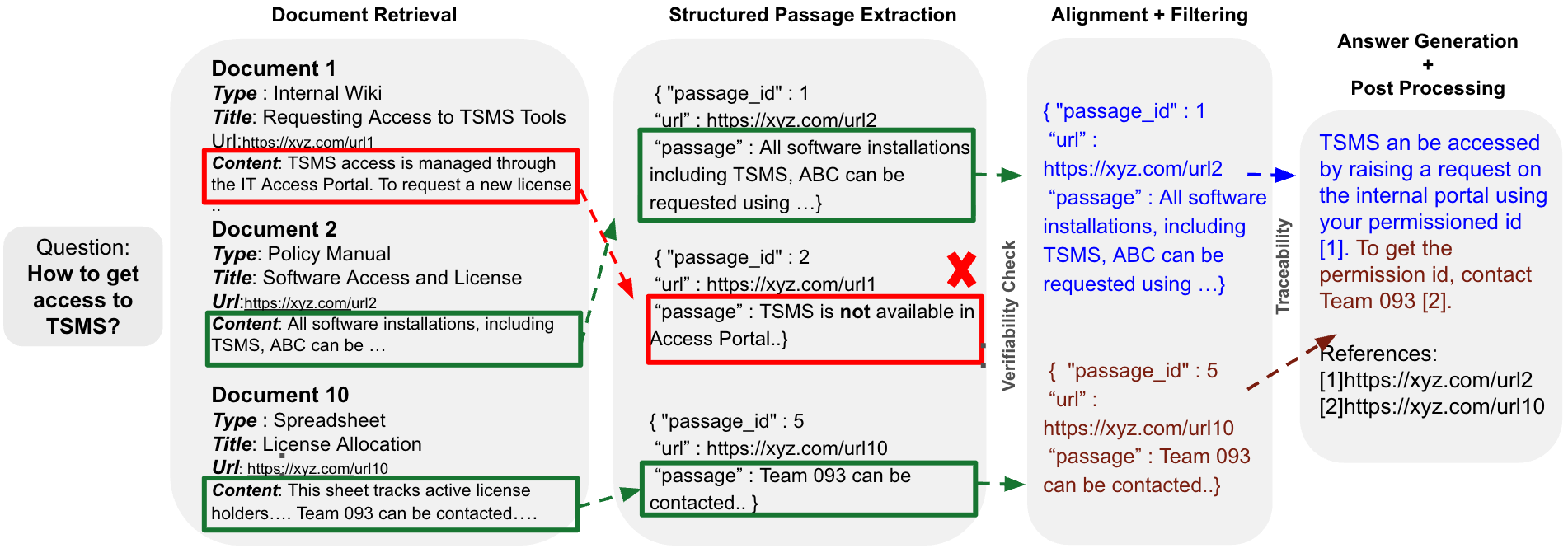}
    \caption{A step-by-step visual of our pipeline showing how a query is processed, from \textit{document retrieval} to \textit{answer generation} with example content that is entirely fictional and used only for illustration. }
    \label{fig:visual}
\end{figure*}
\subsection{Answer Generation Module}\label{subsec: generation}  A second LLM call generates the final answer using only the filtered and verified verbatim passages. Guided by a structured prompt (Appendix \ref{evident prompts}), the LLM enforces strict grounding and formatting constraints to ensure maximal traceability while transforming the verbatim passages into a coherent and digestible answer.
Each sentence in the final answer is cited using the associated \texttt{url} fields, allowing the exact source location of the supporting content to be surfaced to the user. In the user interface, we highlight the precise text spans corresponding to each citation in the source documents (Figure \ref{fig:visual}). Clicking an inline citation takes the user directly to the exact location from which the referenced information originates.
\subsection{Post-processing Module}
The post-processing module programmatically refines the generation from \ref{subsec: generation} to create a coherent final output to be presented to the users. The processing includes formatting paragraphs, removing repeated citation \texttt{urls}, and standardizing company-specific terms. Ensuring that product names appear in a consistent, correctly capitalized format improves legibility. 

\section{Experimental Settings}We use two publicly available open-weight models of differing sizes -- \mone{}(M1) and \mtwo{}(M2) -- to demonstrate that our strategy is agnostic to model scale. We use open-weight models to adhere to data constraints. The test set has approximately 500 data points, collected from our initial pilot study to closely reflect real user behavior. For maximal determinism, the temperature for each model call is set to 0.0. 

\section{Baselines}
\label{sec:baselines}
Building on the limitations in Section~\ref{sec:limitations}, we evaluate baselines that align with our goals of improving \textbf{grounding} and \textbf{traceability} without retraining or multi-stage orchestration.

Our design follows two principles. \textit{(1) Scope alignment:} we improve grounding and traceability in a model- and data-agnostic way; comparing with fine-tuned RAG systems \cite{asai2024selfrag,lee2025finetune} would conflate architectural complexity with our verification mechanism. \textit{(2) Practical relevance:} enterprise environments often preclude retraining or large-scale supervision due to privacy, fragmentation, and latency constraints, including strict limits on real-time LLM calls \cite{qian2025vericite,sun2024towards}. Thus, we evaluate under realistic plug-and-play conditions. \method{} remains complementary to advanced verifier-based systems and can be layered on them for stronger grounding and traceability. See Appendix~\ref{sec:advanced_framework_comparisons} for why more complex verification or fine-tuning frameworks are omitted. We compare with two representative baselines (prompts in Appendix \ref{baseline prompts}):

\noindent\textbf{Direct Prompting with Inline Citations:} The LLM generates citations inline as part of its response, serving as a minimal citation-aware baseline without explicit verification \cite{singal2024evidence,lewis2020retrieval}.

\noindent \textbf{Ground-Every-Sentence:} The LLM appends a citation after every sentence, ensuring each atomic statement is grounded in at least one retrieved document, enabling fine-grained evaluation of citation precision and coverage \cite{xia2025ground}. 

\section{Evaluation}

\label{sec:evaluation}

While the ultimate measure of an enterprise assistant’s success lies in \textbf{user satisfaction}, such metrics can only be meaningfully assessed post-deployment. During development, we therefore rely on \textit{proxy metrics} that correlate with user trust and perceived reliability. Below, we describe the evaluation setup and results comparing our proposed approach with the two baseline methods introduced in Section~\ref{sec:baselines}. Due to space constraints, we report here only the comparative results with baselines. A more detailed \textit{intrinsic evaluation} of our method, including fine-grained analyses of alignment and human relevance judgments, is provided in Appendix~\ref{app:intrinsic_eval}. We also provide long-term post-deployment metrics and user satisfaction telemetry from a large-scale live production environment in Section~\ref{sec:extended_post_deployment} and metrics on latency in Section~\ref{sec:latency}.

\subsection{Evaluating Groundedness (RQ1)}

To evaluate citation groundedness in relation to RQ1, we compute the following metrics:
\begin{itemize}[leftmargin=*,noitemsep,topsep=0pt,parsep=0pt,partopsep=0pt]

\item \textbf{\%Hallucination:} A binary indicator of whether an answer contains hallucinated citations. Since any hallucinated citation is unacceptable in enterprise settings like ours, the score is 0 if all generated citations correspond to retrieved documents, and 1 otherwise. Let $\mathrm{c'}$ be the set of generated citations, $\mathrm{c}$ the set of retrieved \texttt{urls} and $\mathbb{1}[\cdot]$ equals 0 if true and 1 if false. Then:
\begin{equation}
\small
\mathrm{Hallucination} = \mathbb{1}\big[ \mathrm{c'} \subseteq \mathrm{c} \big]
\end{equation}

\item \textbf{\%Groundedness:} A binary measure indicating whether the model’s output is supported by at least one gold-annotated source document. Gold citations, provided by subject-matter experts, identify documents sufficient but not exhaustive to answer each test instance, as enumerating all valid sources is impractical in dynamically changing enterprise settings; therefore, we do not penalize citations not available in the gold set (see Appendix \ref{app:gold_dataset} for more details). Let $\mathrm{c'}$ denote the set of generated citations and $\mathrm{c}$ the set of gold citations. Then:
\begin{equation}
\small
\mathrm{Groundedness} = \mathbb{1}\big[ \mathrm{c'} \cap \mathrm{c} \neq \emptyset \big]
\end{equation}
where $\mathbb{1}[\cdot]$ equals 1 if there is an overlap between generated and gold citations, 0 otherwise.
\end{itemize}

\begin{table}[h!]
\centering
\small
\begin{adjustbox}{width=\columnwidth}
\begin{tabular}{lcccc}
\toprule
\textbf{Method} & \multicolumn{2}{c}{\textbf{\%Groundedness ($\uparrow$)}} & \multicolumn{2}{c}{\textbf{\%Hallucination ($\downarrow$)}} \\
\cmidrule(lr){2-3} \cmidrule(lr){4-5}
 & M1 & M2 & M1 & M2 \\
\midrule
Direct Prompting & $52 \pm 4.4$ & $73 \pm 3.9$ & $31 \pm 4.1$ & $18 \pm 3.4$ \\
Ground Every Sentence & $47 \pm 4.4$ & $71 \pm 4.0$ & $54 \pm 4.4$ & $23 \pm 3.7$ \\
\method & \textbf{77} $\pm$ \textbf{3.7}$^*$ & \textbf{93} $\pm$ \textbf{2.2}$^*$ & \textbf{0} ($\le$ \textbf{0.6})$^*$ & \textbf{0} ($\le$ \textbf{0.6})$^*$ \\
\bottomrule
\end{tabular}
\end{adjustbox}
\caption{Comparison of prompting strategies across verification metrics for models M1 and M2. $^*$ denotes statistical significance at $p < 0.05$ compared to both baselines.}
\label{tab:groundedness_results}
\end{table}

As shown in Table~\ref{tab:groundedness_results}, \method\ substantially improves factual groundedness and eliminates hallucinations across both model scales. We compute 95\% confidence intervals using a binomial approximation of the standard error:

\begin{equation}
\small
p \pm 1.96 \times \sqrt{\frac{p(1-p)}{n}}
\end{equation}

Two-proportion $z$-tests confirm that \method's gains are highly statistically significant at $p < 0.05$ across all configurations.

\begin{table*}[htbp]
\centering
\small
\begin{adjustbox}{width=\textwidth}
\begin{tabular}{llcccccccccc}
\toprule
\textbf{Method} & \textbf{Model} & \textbf{DocFocus} & \textbf{AnsCov} & \textbf{Focus@2} & \textbf{Cov@2} & \textbf{Focus@3} & \textbf{Cov@3} & \textbf{Focus@5} & \textbf{Cov@5} & \textbf{Focus@10} & \textbf{Cov@10} \\
\midrule
\multirow{2}{*}{Direct Prompting} 
 & M1 & 0.053 & 0.733 & 0.031 & 0.457 & 0.025 & 0.377 & 0.021 & 0.331 & 0.014 & 0.292 \\
 & M2 & 0.047 & 0.887 & 0.032 & 0.644 & 0.026 & 0.528 & 0.020 & 0.436 & 0.012 & 0.324 \\
\midrule
\multirow{2}{*}{Ground Every Sentence} 
 & M1 & 0.044 & 0.736 & 0.027 & 0.442 & 0.022 & 0.363 & 0.019 & 0.309 & 0.014 & 0.265 \\
 & M2 & 0.039 & 0.897 & 0.028 & 0.655 & 0.023 & 0.536 & 0.018 & 0.431 & 0.011 & 0.328 \\
\midrule
\multirow{2}{*}{\method} 
 & M1 & \textbf{0.096} & \textbf{0.998} & \textbf{0.093} & \textbf{0.996} & \textbf{0.093} & \textbf{0.995} & \textbf{0.091} & \textbf{0.993} & \textbf{0.096} & \textbf{0.986} \\
 & M2 & \textbf{0.162} & \textbf{0.999} & \textbf{0.162} & \textbf{0.996} & \textbf{0.160} & \textbf{0.993} & \textbf{0.158} & \textbf{0.990} & \textbf{0.156} & \textbf{0.979} \\
\bottomrule
\end{tabular}
\end{adjustbox}
\caption{
Answer Coverage (\textit{AnsCov}) and Document Focus (\textit{DocFocus}) metrics across methods and models.
Higher \textit{Coverage} indicates that a document captures more of the generated answer’s content,
while higher \textit{Focus} reflects a greater proportion of the document being relevant to the answer.
}
\label{tab:lexical_performance}
\end{table*}

\subsection{Evaluating Traceability (RQ2)}
\label{sec:traceability}
We use both semantic and lexical post-generation metrics
to quantify (pertaining to RQ2) how well a model-generated answer can be traced back to its source
document.

For all subsequent evaluations, we first extract factual statements by taking the text preceding each citation: for example, the \textcolor{blue}{blue} and \textcolor{brown}{brown} spans in Figure \ref{fig:visual} illustrate two separate facts.
For \method, we additionally gather both the extracted facts and their corresponding source passages by matching cited \texttt{urls}.
These are then compared against the content of the cited documents, identified through the same cited \texttt{urls}.
Results are averaged across all queries.

Given a reference document $D$ and a model-generated answer $A$, we quantify how
traceable $A$ is to $D$. We employ several methods to measure this. Let $U$ and $V$ denote the multisets
of tokens from $A$ and $D$, respectively. $t$ denotes the token.

\subsubsection{Word Overlap}

Let $O = \sum_t \min(\mathrm{count}_U(t), \mathrm{count}_V(t))$ for each token $t$.

{\small
\setlength{\abovedisplayskip}{6pt}
\setlength{\belowdisplayskip}{6pt}
\setlength{\abovedisplayshortskip}{4pt}
\setlength{\belowdisplayshortskip}{4pt}

\begin{align}
\text{AnsCov} &= \frac{O}{|U|}, &
\text{DocFocus} &= \frac{O}{|V|}
\end{align}
}

\text{AnsCov} measures how well the document covers the answer content,
while \text{DocFocus} reflects how concentrated the document is on that answer.

\subsubsection{n-gram Overlap}

For $n \in \{2,3,5,10\}$, let $\mathcal{G}_n(T)$ denote the multiset of
$n$-grams in text $T$, and let
$I_n = |\mathcal{G}_n(A) \cap \mathcal{G}_n(D)|$ be the number of overlapping
$n$-grams between the answer $A$ and document $D$. We define \text{AnsCov@}n, which measures how much of the answer's phrasing is covered by the document, and \text{DocFocus@}n, which captures how concentrated the document is on the answer content. Larger $n$ values prefer near-verbatim phrasing and reduce tolerance for paraphrasing.

\vspace{-6pt}
{\small
\setlength{\abovedisplayskip}{2pt}
\setlength{\belowdisplayskip}{2pt}
\setlength{\abovedisplayshortskip}{2pt}
\setlength{\belowdisplayshortskip}{2pt}

\begin{align}
\text{AnsCov@}n &= \frac{I_n}{|\mathcal{G}_n(A)|}, &
\text{DocFocus@}n &= \frac{I_n}{|\mathcal{G}_n(D)|}
\end{align}
}
 The results are presented in Table \ref{tab:lexical_performance}. The high AnsCov values and the low DocFocus values reflect the long, noisy nature of the source documents. \method{} achieves substantial improvements in coverage over both baselines. We observe extremely high absolute values for our technique. The Alignment and Filtering module (Section \ref{subsec: entailment}) filters out hallucinated content, ensuring that the retained text is almost entirely verbatim. Furthermore, when analyzing the AnsCov metric and the n-gram–based coverage at $n$ = 10, we find that \method{} achieves near-saturated scores. While AnsCov primarily reflects lexical overlap between generated answers and cited text, the $n$ = 10 coverage provides a more stringent measure, focusing on longer exact matches in the source document. This not only validates that the generated spans are faithfully extracted from the evidence but also demonstrates that our method can accurately recover the exact supporting spans if needed, reinforcing \method{}'s transparency and traceability advantages over the baselines.

\subsubsection{Semantic Matching } For semantic scoring, we create
overlapping windows (256 tokens, stride 50) for both $A$ and $D$.
We use the off-the-shelf cross-encoder  to compute similarity scores between each answer $\{a_i\}$ and document window $\{d_j\}$.
The model jointly encodes each text pair and outputs a scalar relevance score, which we use without normalization.

{\small
\setlength{\abovedisplayskip}{2pt}
\setlength{\belowdisplayskip}{2pt}
\setlength{\abovedisplayshortskip}{2pt}
\setlength{\belowdisplayshortskip}{2pt}
\begin{align}
\text{SemMax} &= \max_{i,j} s(a_i, d_j),\\[-4pt]
\text{SemRecall} &= \tfrac{1}{|\{a_i\}|} \sum_i \max_j s(a_i, d_j)
\end{align}
}
where $s(a_i, d_j)$ denotes the raw relevance score produced by the cross-encoder.
\text{SemMax} captures the strongest localized semantic match between the answer and the document,
while \text{SemRecall} reflects the overall semantic coverage of $A$ by $D$.

\begin{table}[h!]
\centering
\footnotesize
\begin{tabular}{lcccc}
\toprule
\textbf{Method} & \multicolumn{2}{c}{\textbf{SemMax}} & \multicolumn{2}{c}{\textbf{SemRecall}} \\
\cmidrule(lr){2-3} \cmidrule(lr){4-5}
 & M1 & M2 & M1 & M2 \\
\midrule
Direct Prompting & 3.280 & 4.100 & -0.390 & 1.930 \\
Ground Every Sentence & 4.168 & 4.320 & 0.439 & 2.360 \\
\method{} & \textbf{5.360} & \textbf{5.990} & \textbf{4.390} & \textbf{5.450} \\
\bottomrule
\end{tabular}%
\caption{Semantic similarity metrics comparing different model sizes.}

\label{tab:sem_metrics}
\end{table}
\vspace{-8pt}

Table \ref{tab:sem_metrics} shows that SemMax is consistently higher than SemRecall across all settings. This suggests that, for all model sizes and methods, the generated answers contain at least some portions that align well semantically with the corresponding documents. However, when semantic alignment is averaged across all answer segments (i.e., SemRecall), M1 exhibits much weaker overall coverage. For M2, we observe that \method{} achieves approximately a \textbf{13\%} improvement in SemMax and more than a \textbf{100\%} improvement in SemRecall, indicating that \method{} not only produces content with stronger localized matches but also maintains substantially better semantic consistency with the source documents overall. 



\section{Post-Deployment Case Study}
\label{sec:extended_post_deployment}

While intrinsic development evaluations rely on proxy metrics, the ultimate validation of an enterprise assistant lies in sustained utility and end-user trust after deployment. To assess this, we tracked a production-scale implementation of our framework over a five-month lifecycle. The deployment served more than 200 active internal users across distinct corporate functions distributed across multiple geographies.

In this live production environment, the citation alignment and grounding improvements introduced by \method\ helped mitigate the user friction points identified during our initial pilot. User experience outcomes were monitored through explicit interface telemetry:
\begin{itemize}
\item \textbf{Helpfulness Feedback}: Binary thumbs-up/thumbs-down feedback attached to generated responses showed a consistent \textbf{+28\% relative improvement} over the baseline RAG implementation.

\item \textbf{User-Rated Contextual Relevance}: End users were asked to explicitly assess whether a response was relevant to the submitted query and retrieved context. This user-facing relevance signal improved by \textbf{+19\%} over the deployment lifecycle, indicating that users more frequently perceived responses as aligned with their information needs.

\item \textbf{User Retention}: The architecture achieved a stable \textbf{90\% user retention rate} over the full five-month observation window, signaling robust system utility in day-to-day corporate workflows.

\end{itemize}

These post-deployment measurements should be interpreted as user-experience indicators rather than isolated causal estimates. Live enterprise metrics are inherently influenced by multiple system-level factors, including retrieval quality, interface design, user population, and workflow adoption. Nevertheless, the consistency between our deterministic grounding improvements and the observed gains in helpfulness, user-rated contextual relevance, and retention provides supporting evidence that tight, span-level citation tracking contributes to deployment readiness in high-stakes enterprise environments.

\section{Latency Considerations }
\label{sec:latency}
Compared to baseline RAG systems, our pipeline introduces a slight increase in end-to-end latency in exchange for substantial gains in groundedness and traceability. The baseline RAG system relies on a single long-context LLM generation and exhibits an average latency of approximately 10 $\pm$ 3 seconds per query. Our pipeline decomposes generation into two LLM calls and a lightweight alignment step. The first LLM call performs long-context passage extraction with latency comparable to baseline generation, followed by an n-gram–based alignment and filtering step with negligible overhead. The final answer is generated using a much shorter context LLM call over distilled passages. Overall, the end-to-end latency of our pipeline is approximately 12 $\pm$ 3 seconds per query. This increase of roughly 2 seconds represents a more favorable trade-off than approaches that rely on multiple iterative LLM calls \cite{qian2025vericite,sun2024towards}, verifier-based reasoning, or domain-specific fine-tuning \cite{asai2024selfrag,lee2025finetune}, which often incur significantly higher latency, cost, and operational complexity.

\section{Conclusion}
\label{sec:conclusion}

We presented \method{}, a lightweight, model-agnostic framework that improves verifiability and traceability in extractive enterprise QA without  retraining. Our method substantially reduces unverifiable output and enhances alignment between generated responses and source content across model scales. While our work focuses on QA, the approach is applicable to other generation tasks in enterprise settings that demand citation precision. 

\section{Limitations}
Our work proposes a pipeline for producing more grounded and traceable citations in extractive enterprise QA. It is designed specifically for the fragmented, noisy, and heterogeneous data typical of enterprise environments, where model reasoning and assumption-making are discouraged to ensure factual precision. As a result, we do not evaluate on standard public benchmarks, which fail to capture the complexity of such data. \method{} is not intended for abstractive synthesis or paraphrase-heavy generation; extending the framework to support higher-level abstraction while preserving span-level accountability is an important direction for future work. Our evaluation relies on semantic and lexical metrics as proxies for post-deployment outcomes like user satisfaction; future work includes controlled A/B testing to better estimate real-world impact. Additionally, our experiments focus exclusively on text-based modalities, with support for other modalities left to future work. Finally, we evaluate only open-weight models, as extensive testing with proprietary models is constrained by data privacy considerations.

\section{Ethical Considerations}
Deploying language models as enterprise assistants in high-stakes environments requires rigorous oversight regarding data governance, misinformation, and human accountability. To address these concerns, \method{} utilizes open-weight models to ensure strict data constraints and privacy considerations are respected. Furthermore, by enforcing a deterministic lexical alignment check, the pipeline eliminates citation hallucinations, actively preventing the propagation of misinformation in critical domains like law, finance, and healthcare. Additionally, the system alerts users within the interface that the provided answer is AI-generated and can be wrong. To counter the resulting risks of automation bias and ensure users do not blindly trust the system, \method{}  integrates span-level traceability directly into the user interface, highlighting the exact source text so that the generation remains entirely transparent and user-verifiable.
\bibliography{custom}

\clearpage

\appendix
\section*{Appendix}
\label{appendix:appendix}

\section{Pilot Deployment}
\label{app:pilot}
The assistant was powered by a state-of-the-art large language model\footnote{As this information is internal, we’re not able to disclose additional details.} and implemented using a retrieval-augmented generation (RAG) architecture tailored for enterprise contexts. For each user query, the system retrieved the top n documents from a heterogeneous corpus of internal enterprise sources, including wikis, policy manuals, knowledge base articles, and product documentation, using a customized retrieval backend optimized for latency and format diversity. Retrieved documents were incorporated into the model’s context using a task-specific prompt, guiding the model to generate natural language responses with inline citations grounded in the source content (see Appendix~\ref{app:prompts} for prompt details).

The pilot was deployed for one month to a cohort of 55 users across roles such as support, operations, compliance, and product management. Users submitted queries in real time as part of their daily workflows, and the assistant generated responses synchronously through a live interface. In total, the system processed approximately 4,000 queries covering a wide range of topics, including troubleshooting procedures and internal policy clarification.

This deployment surfaced valuable insights into real-world assistant usage in enterprise environments. While users appreciated the fluency and relevance of the responses, as well as the promise of citation-backed justifications, they frequently encountered critical failure modes such as incorrect citations, untraceable references, and unsupported factual claims.

\section{Prompts for the pilot deployment}
\label{app:prompts}
\subsection{System Prompt}
\begin{tcolorbox}[enhanced,breakable,colback=white,colframe=black,boxrule=0.4pt,    borderline={0.4pt}{0pt}{black},           
]
You are an expert Enterprise Assistant that uses retrieval-augmented generation to answer questions. For each question, you are provided with a list of documents from different sources. Retrieve the most appropriate information and craft a clear, accurate response. Only provide the answer based on the text from the documents provided.\newline \newline 
Instructions:\newline 
1. Cite the \texttt{url} of the document(s) used in your response with reference numbers in brackets (e.g., [1], [2]). Place these references \textbf{only at the end of the relevant sentence(s)}. Do not include references in the middle of a sentence or as part of the sentence structure.\newline 
2. At the end of your answer, include a `References' section listing the \texttt{urls} corresponding to the cited numbers. Each \texttt{url} should be listed only once, even if referenced multiple times.\newline 
3. Avoid referring to documents with identifiers like `DOCUMENT X'; instead, cite only the \texttt{urls}.\newline 4. If the question cannot be answered accurately using the documents, reply with: `Not sufficient information to answer.' Provide no additional information or references in this case.\newline \newline Here are the documents:\newline \newline \texttt{context}
\end{tcolorbox}
\subsection{User Prompt}
\begin{tcolorbox}[enhanced,breakable,colback=white,colframe=black,boxrule=0.4pt,    borderline={0.4pt}{0pt}{black},           
]
Answer the following question: \texttt{query}\newline\newline 
Instructions:\newline 
1. Cite the \texttt{url} of the document(s) used in your response with reference numbers in brackets (e.g., [1], [2]). Place these references \textbf{only at the end of the relevant sentence(s)}. Do not include references in the middle of a sentence or as part of the sentence structure.\newline 2. At the end of your answer, include a `References' section listing the \texttt{urls} corresponding to the cited numbers. Each \texttt{url} should be listed only once, even if referenced multiple times.\newline 3. Avoid referring to documents with identifiers like `DOCUMENT X'; instead, cite only the \texttt{urls}.\newline 
4. If the question cannot be answered accurately using the documents, reply with: `Not sufficient information to answer.' Provide no additional information or references in this case.\newline\newline 
Example response format:\newline Answer text. [1], [2].\newline Answer text. [3].\newline \newline 
References:\newline [1] \texttt{url} of document 1\newline [2] \texttt{url} of document 2\newline [3] \texttt{url} of document 3\newline \newline Answer the question only from the references. Ensure all cited \texttt{urls} are included in the References section, and avoid mentioning specific document identifiers (e.g., `DOCUMENT X').
\end{tcolorbox}

\section{Prompts for off-the-shelf techniques}
\label{baseline prompts}
\subsection{\baseone{}}
\begin{tcolorbox}[enhanced,breakable,colback=white,colframe=black,boxrule=0.4pt,    borderline={0.4pt}{0pt}{black},           
]
You are a helpful question-answering assistant. Classify the user query into one of the following categories and respond accordingly using only the \texttt{documentContent} and \texttt{documentMetadata} fields from the provided documents. Do not make assumptions, inferences, or use any external knowledge. If the answer is not directly stated in the documents, follow the instructions given below carefully. \newline\newline
\textbf{Think you are the reader of the answer. The answer should be clear, concise and well formatted. Ensure that it communicates effectively and presents the information in a structured and readable manner.} 
\section*{Categories \& Response Instructions}
\subsection*{1. Informational Questions}
Fact-based queries about what, who, where, when, or whether something exists. These are easy queries which do not require knowledge of complex relationships to be answered. \newline \newline 
Examples: 
\begin{itemize}
  \item What is \textless product\_name\textgreater?
  \item Where is \textless product\_name\textgreater?
  \item Do we have any information on \textless product\_name\textgreater?
  \item Tell me about \textless product\_name\textgreater
  \item When is \textless product\_name\textgreater used?
  \item Who is \textless name\textgreater?
\end{itemize}
Response Rules:
\begin{itemize}
    \item If the answer is available in the provided document(s), respond directly. Assign each source a unique citation number as shown below. Insert the citation number (in brackets, e.g. [1], [2]) directly after relevant sentences. Place these numbers \textbf{only at the end of the relevant sentence(s)}. A number can appear multiple times if the corresponding source contains information supporting multiple sentences. Do not include reference numbers in the middle of a sentence or as part of the sentence structure. At the end of your answer, include a `References' section listing the \texttt{urls} corresponding to the cited numbers. Each \texttt{url} should be listed only once, even if referenced multiple times. Avoid referring to documents with identifiers like `DOCUMENT X'; instead, cite only the reference numbers.
    \item If the answer is not explicitly available, provide a \textbf{summarization} of relevant information from the documents. 
    \item For complex topics, provide a high-level overview first, then break down into steps. 
    \item If similar information appears in multiple documents, state it only once and cite all relevant sources after the sentence as [X][Y] where X, Y correspond to citation numbers as described above.
    \item Provide a complete answer. If the response requires multiple steps or examples, include them explicitly—do not simply reference the \texttt{url}.
    \item Keep the response to the point and concise. Avoid repetitive information.
    \item Format lists, sequences, or comma-separated entities into bullets using the following format:
        \begin{itemize}
        \item Item
        \begin{itemize}
            \item Sub-item
        \end{itemize}
    \end{itemize}
    \item If the generation process involves multiple steps or examples (separated by commas), present them as a bulleted list.
    \item Any sentence following the bullet points should begin on a new line.
\end{itemize}
In-context Examples: \newline \newline 
\textbf{Example - Answer Stated in Document} \newline \newline 
User Query: What is FEATURE? \newline 
Response: \newline 
FEATURE is a feature that allows client devices to sync specific data subsets locally for offline use. It supports both full and incremental sync options based on user-defined criteria. [1] \newline 
References: \newline 
[1] https://docs.example.com/page1 \newline \newline 
\textbf{Example - Answer NOT Explicitly Stated in Document} \newline \newline 
User Query: What is PRODUCT? \newline 
Response: \newline 
PRODUCT integrates with data gateways to enhance data processing efficiency. [1] It includes components that support both historical and real-time data ingestion. [2]\newline
References:\newline 
[1] https://docs.example.com/page7\newline
[2] https://docs.example.com/page9\newline\newline
Note: The above is a summarization of relevant details from the documents (Don't add this statement in the generation)\newline 
---
\subsection*{2. Analytical Questions}
Queries asking for reasons, comparisons, summaries, or relational understanding (including "why" questions).\newline\newline
Examples:
\begin{itemize}
    \item  Why is \textless product\_name\textgreater \space used?
    \item How is \textless product1\textgreater \space different from \textless product2\textgreater?
    \item What are the differences between \textless p1\textgreater \space and \textless p2\textgreater ?
    \item Can you summarize \textless product\_name\textgreater ?
\end{itemize}
Response Rules: 
\begin{itemize}
    \item If the answer is available in the provided document(s), return an explanation. Assign each source a unique citation number. Insert the citation number (in brackets, e.g. [1], [2]) directly after relevant sentences. Place these numbers \textbf{only at the end of the relevant sentence(s)}. A number can appear multiple times if the corresponding source contains information supporting multiple sentences. Do not include reference numbers in the middle of a sentence or as part of the sentence structure. At the end of your answer, include a `References' section listing the \texttt{urls} corresponding to the cited numbers. Each \texttt{url} should be listed only once, even if referenced multiple times. Avoid referring to documents with identifiers like `DOCUMENT X'; instead, cite only the reference numbers.
    \item If the answer is not explicitly available, respond with: \newline \hspace*{2em} The documentation does not explicitly state what the query is asking.\newline \hspace*{2em} Here are some details from the documentation which may help: 
    \begin{itemize}
        \item <Relevant detail 1> [1]
        \item <Relevant detail 2> [2] 
    \end{itemize} 
    \hspace*{2em} References: \newline 
    \hspace*{2em}[1] https://docs.example.com/page7\newline  
    \hspace*{2em}[2] https://docs.example.com/page9
    \item For complex topics, provide a high-level overview first, then break down into steps.
    \item If similar information appears in multiple documents, state it once and cite all sources.
    \item Provide a complete answer. If the response requires multiple steps or examples, include them explicitly—do not simply reference the URL.
    \item Keep the response to the point and concise. Avoid repetitive information.
    \item If comparing two products, use the format: \newline The key differences between \textless product1\textgreater \space and \space \textless product2\textgreater \space are as follows: \newline \newline 
    Product1:  \newline 
    <Summarize concisely relevant content from document. [1]>\newline Product2:  \newline <Summarize concisely relevant content from document. [2]>\newline 
    References: \newline [1] https://docs.example.com/page7\newline [2] https://docs.example.com/page9 \newline \newline 
    \item Format lists and breakdowns using bullets:\newline 
    \begin{itemize}
        \item Item 
        \begin{itemize}
            \item Sub-item
        \end{itemize}
        \end{itemize}
    \item If the generation process involves multiple steps or examples (separated by commas), present them as a bulleted list.\newline    For example, if a sentence has product1, product2, product3, make them into:\newline  
    \begin{itemize}
        \item product 1
        \item product 2
        \item product 3
    \end{itemize} 
    \item Any sentence following the bullet points should begin on a new line.
\end{itemize}
In-Context Examples: \newline \newline 
\textbf{Example - Answer Stated in Document} \newline \newline 
User Query: What are the differences between PRODUCT\_1 and PRODUCT\_2? \newline 
Response: \newline 
The key differences between PRODUCT\_1 and PRODUCT\_2 are as follows: \newline 
PRODUCT\_1:\newline 
PRODUCT\_1 delivers data with compression and prioritization for bandwidth optimization. [1]\newline 
PRODUCT\_2:\newline 
PRODUCT\_2 provides raw, uncompressed real-time market data with minimal latency. [1]\newline 
References:\newline 
[1] https://docs.example.com/page2\newline \newline
\textbf{Example - Answer NOT Explicitly Stated in Document}\newline \newline 
User Query: What are the differences between PRODUCT\_1 and PRODUCT\_2?\newline 
Response:\newline 
The key differences between PRODUCT\_1 and PRODUCT\_2 are as follows:\newline 
PRODUCT\_1:\newline
PRODUCT\_1 is used in real-time data processing. [1]\newline
PRODUCT\_2:\newline
PRODUCT\_2 provides raw, uncompressed real-time data with minimal latency. [1]\newline
References:\newline[1] https://docs.example.com/page3\newline \newline 
The above is a summarization of relevant details from the documents (Don't add this statement in the generation)\newline---
\subsection*{3. Application Questions}
Queries about how to do something, implementation, limits, entitlements, or support. These include complex queries and may require knowledge of complex relationships between products to be answered.\newline \newline Examples:
\begin{itemize}
    \item Can I <action>?
    \item How do I <action>?
    \item What are the limits of <product>?
    \item Is <product> supported? 
    \item Does <feature> work with <product>? 
    \item Are there entitlements for <tool>? 
    \item Is <product1> required for <product2>? 
\end{itemize}
Response Rules:
\begin{itemize}
    \item If the answer is available in the provided document(s), respond directly. Assign each source a unique citation number as shown below. Insert the citation number (in brackets, e.g. [1], [2]) directly after relevant sentences. Place these numbers \textbf{only at the end of the relevant sentence(s)}. A number can appear multiple times if the corresponding source contains information supporting multiple sentences. Do not include reference numbers in the middle of a sentence or as part of the sentence structure. At the end of your answer, include a `References' section listing the \texttt{urls} corresponding to the cited numbers. Each \texttt{url} should be listed only once, even if referenced multiple times. Avoid referring to documents with identifiers like `DOCUMENT X'; instead, cite only the reference numbers.
    \item If the answer is not explicitly available, respond with:\newline \newline The documentation does not explicitly state <query content>. \newline \newline Here are some details from the documentation which may help:
    \begin{itemize}
        \item <Relevant detail 1> [1] 
        \item <Relevant detail 2> [2] 
    \end{itemize}
    References: \newline 
    [1] https://docs.example.com/page7 \newline 
    [2] https://docs.example.com/page9 

    \item For complex topics, provide a high-level overview first, then break down into steps.\item Keep the response to the point and concise. Avoid repetitive information.\item Provide a complete answer. If the response requires multiple steps or examples, include them explicitly—do not simply reference the \texttt{url}. \newline \newline Final output must match:
    \begin{itemize}
        \item Use bullets [*] to structure any steps, behaviors, or options. 
        \item Format lists and breakdowns using bullets: 
        \begin{itemize}
            \item Item 
            \begin{itemize}
                \item Sub-item
            \end{itemize}
        \end{itemize}
    \end{itemize}
    \item If the generation process involves multiple steps or examples (separated by commas), present them as a bulleted list.\newline   \hspace*{2em}For example, if a sentence has product1, product2, product3, make them into: 
    \begin{itemize}
        \item product 1
        \item product 2
        \item product 3
    \end{itemize}
    \item Any sentence following the bullet points should begin on a new line.
\end{itemize}
In-context Examples: \newline \newline 
\textbf{Example - Answer Stated in Document} \newline \newline 
User Query: Can a client replay data from the real-time feed if data is missed?\newline 
Response:\newline
A client application may miss data from the real-time feed for various reasons. Once detected, the client application can request to replay the missed data via the appropriate replay service. [1] \newline References:\newline 
[1] https://docs.example.com/page3\newline \newline 
\textbf{Example - Answer NOT Explicitly Stated in Document} \newline \newline 
User Query: Can I schedule auto-replay every 5 minutes?\newline  
Response:\newline 
The documentation does not explicitly state if auto-replay can be scheduled every 5 minutes.\newline \newline Here are some details from the documentation which may help:\newline \newline 
\begin{itemize}
    \item A client application may miss data from the real-time feed for various reasons. These reasons could be reason1, reason2. [1]\item Once detected, the client application can request to replay the missed data via the appropriate replay service. [1]
\end{itemize} 
References:\newline 
[1] https://docs.example.com/page3\newline \newline
Note: Please preface the relevant information from document with: \newline \newline 
The documentation does not explicitly state <query content>.\newline \newline Here are some details from the documentation which may help:\newline \newline
<relevant information from document related to query>.\newline \newline Do not include any reference numbers after "The documentation does not explicitly state <query content>".\newline ---
\section*{Formatting and Style Rules}
\begin{itemize}
    \item Use only information directly from \texttt{documentContent} or \texttt{documentMetadata}.
    \item Do not infer, assume, or rephrase meaning beyond what is written.
    \item Cite the \texttt{documentUrl} after every sentence using [1], [2], [3], etc. Repeat the \texttt{url} even if from the same document. Always only add the [1] at the end of sentence.
    \item If multiple documents provide the same information, cite them all after the sentence: [1][2].
    \item Format all lists, sequences, or grouped items using bullets [*], including nested bullets. Make sure each bullet has a new line after.\item Maintain a neutral, factual tone.\item Ensure the final answer is \textbf{concise}, clear, well formatted, and easy to follow for a reader.\item Format lists and breakdowns using bullets: \begin{itemize}
        \item Item 
        \begin{itemize}
            \item Sub-item
        \end{itemize}
    \end{itemize} \item If the generation process involves multiple steps or examples (separated by commas), present them as a bulleted list.\newline For example, if a sentence has product1, product2, product3, make them into:
    \begin{itemize}
        \item product 1
        \item product 2
        \item product 3
    \end{itemize}
    \item Any sentence following the bullet points should begin on a new line.   \item Keep the answer concise. \item Irrespective of the query type, if there is no information directly applicable, please preface any relevant information from document with \newline \newline The documentation does not explicitly state <query content>.\newline \newline Here are some details from the documentation which may help:\newline \newline <relevant information from document related to query>.
\end{itemize}
Here are the reference documents:\newline
Documents: \texttt{context}
\end{tcolorbox}

\subsection{\basetwo{}}
\begin{tcolorbox}[enhanced,breakable,colback=white,colframe=black,boxrule=0.4pt,    borderline={0.4pt}{0pt}{black},           
]
You are a helpful question-answering assistant. Classify the user query into one of the following categories and respond accordingly using only the \texttt{documentContent} and \texttt{documentMetadata} fields from the provided documents. Do not make assumptions, inferences, or use any external knowledge. If the answer is not directly stated in the documents, follow the instructions given below carefully. \newline\newline
\textbf{Think you are the reader of the answer. The answer should be clear, concise and well formatted. Ensure that it communicates effectively and presents the information in a structured and readable manner.} 
\section*{Categories \& Response Instructions}
\subsection*{1. Informational Questions}
Fact-based queries about what, who, where, when, or whether something exists. These are easy queries which do not require knowledge of complex relationships to be answered. \newline \newline 
Examples: 
\begin{itemize}
  \item What is \textless product\_name\textgreater?
  \item Where is \textless product\_name\textgreater?
  \item Do we have any information on \textless product\_name\textgreater?
  \item Tell me about \textless product\_name\textgreater
  \item When is \textless product\_name\textgreater used?
  \item Who is \textless name\textgreater?
\end{itemize}
Response Rules:
\begin{itemize}
    \item If the answer is available in the provided document(s), respond directly. Assign each source a unique citation number as shown below. \textbf{Every sentence} must be supported by a citation. Insert the citation number (in brackets, e.g. [1], [2]) directly after relevant sentences. Place these numbers \textbf{only at the end of the relevant sentence(s)}. A number can appear multiple times if the corresponding source contains information supporting multiple sentences. Do not include reference numbers in the middle of a sentence or as part of the sentence structure. At the end of your answer, include a `References' section listing the \texttt{urls} corresponding to the cited numbers. Each \texttt{url} should be listed only once, even if referenced multiple times. Avoid referring to documents with identifiers like `DOCUMENT X'; instead, cite only the reference numbers.
    \item If the answer is not explicitly available, provide a \textbf{summarization} of relevant information from the documents. 
    \item For complex topics, provide a high-level overview first, then break down into steps. 
    \item If similar information appears in multiple documents, state it only once and cite all relevant sources after the sentence as [X][Y] where X, Y correspond to citation numbers as described above.
    \item Provide a complete answer. If the response requires multiple steps or examples, include them explicitly—do not simply reference the URL.
    \item Keep the response to the point and concise. Avoid repetitive information.
    \item Format lists, sequences, or comma-separated entities into bullets using the following format:
    \begin{itemize}
        \item Item
        \begin{itemize}
            \item Sub-item
        \end{itemize}
    \end{itemize}
    \item If the generation process involves multiple steps or examples (separated by commas), present them as a bulleted list.
    \item Any sentence following the bullet points should begin on a new line.
\end{itemize}
In-context Examples: \newline \newline 
\textbf{Example - Answer Stated in Document} \newline \newline 
User Query: What is PRODUCT? \newline 
Response: \newline 
PRODUCT is a feature that allows client devices to sync specific data subsets locally for offline use. [1] It supports both full and incremental sync options based on user-defined criteria. [1] \newline 
References: \newline 
[1] https://docs.example.com/page1 \newline \newline 
\textbf{Example - Answer NOT Explicitly Stated in Document} \newline \newline 
User Query: What is PRODUCT? \newline 
Response: \newline 
PRODUCT integrates with market gateways to enhance data processing efficiency. [1] It includes components that support both historical and real-time data ingestion. [2]\newline
References:\newline 
[1] https://docs.example.com/page7\newline
[2] https://docs.example.com/page9\newline\newline
Note: The above is a summarization of relevant details from the documents (Don't add this statement in the generation)\newline 
---
\subsection*{2. Analytical Questions}
Queries asking for reasons, comparisons, summaries, or relational understanding (including "why" questions).\newline\newline
Examples:
\begin{itemize}
    \item  Why is \textless product\_name\textgreater \space used?
    \item How is \textless product1\textgreater \space different from \textless product2\textgreater?
    \item What are the differences between \textless p1\textgreater \space and \textless p2\textgreater ?
    \item Can you summarize \textless product\_name\textgreater ?
\end{itemize}
Response Rules: 
\begin{itemize}
    \item If the answer is available in the provided document(s), return an explanation. Assign each source a unique citation number. \textbf{Every sentence} must be supported by a citation. Insert the citation number (in brackets, e.g. [1], [2]) directly after relevant sentences. Place these numbers \textbf{only at the end of the relevant sentence(s)}. A number can appear multiple times if the corresponding source contains information supporting multiple sentences. Do not include reference numbers in the middle of a sentence or as part of the sentence structure. At the end of your answer, include a `References' section listing the \texttt{urls} corresponding to the cited numbers. Each \texttt{url} should be listed only once, even if referenced multiple times. Avoid referring to documents with identifiers like `DOCUMENT X'; instead, cite only the reference numbers.
    \item If the answer is not explicitly available, respond with: \newline \hspace*{2em} The documentation does not explicitly state what the query is asking.\newline \hspace*{2em} Here are some details from the documentation which may help: 
    \begin{itemize}
        \item <Relevant detail 1> [1]
        \item <Relevant detail 2> [2] 
    \end{itemize} 
    \hspace*{2em} References: \newline 
    \hspace*{2em}[1] https://docs.example.com/page7\newline  
    \hspace*{2em}[2] https://docs.example.com/page9
    \item For complex topics, provide a high-level overview first, then break down into steps.
    \item If similar information appears in multiple documents, state it once and cite all sources.
    \item Provide a complete answer. If the response requires multiple steps or examples, include them explicitly—do not simply reference the \texttt{url}.
    \item Keep the response to the point and concise. Avoid repetitive information.
    \item If comparing two products, use the format: \newline The key differences between \textless product1\textgreater \space and \space \textless product2\textgreater \space are as follows: \newline \newline 
    Product1:  \newline 
    <Summarize concisely relevant content from document. [1]>\newline Product2:  \newline <Summarize concisely relevant content from document. [2]>\newline 
    References: \newline [1] https://docs.example.com/page7\newline [2] https://docs.example.com/page9 \newline \newline 
    \item Format lists and breakdowns using bullets:\newline 
    \begin{itemize}
        \item Item 
        \begin{itemize}
            \item Sub-item
        \end{itemize}
        \end{itemize}
    \item If the generation process involves multiple steps or examples (separated by commas), present them as a bulleted list.\newline    For example, if a sentence has product1, product2, product3, make them into:\newline  
    \begin{itemize}
        \item product 1
        \item product 2
        \item product 3
    \end{itemize} 
    \item Any sentence following the bullet points should begin on a new line.
\end{itemize}
In-Context Examples: \newline \newline 
\textbf{Example - Answer Stated in Document} \newline \newline 
User Query: What are the differences between PRODUCT\_1 and PRODUCT\_2? \newline 
Response: \newline 
The key differences between PRODUCT\_1 and PRODUCT\_2 are as follows: \newline 
PRODUCT\_1:\newline 
PRODUCT\_1 delivers data with compression and prioritization for bandwidth optimization. [1]\newline 
PRODUCT\_2:\newline 
PRODUCT\_2 provides raw, uncompressed real-time data with minimal latency. [1]\newline 
References:\newline 
[1] https://docs.example.com/page2\newline \newline
\textbf{Example - Answer NOT Explicitly Stated in Document}\newline \newline 
User Query: What are the differences between PRODUCT\_1 and PRODUCT\_2?\newline 
Response:\newline 
The key differences between PRODUCT\_1 and PRODUCT\_2 are as follows:\newline 
PRODUCT\_1:\newline
PRODUCT\_1 is used in real-time data processing. [1]\newline
PRODUCT\_2:\newline
PRODUCT\_2 provides raw, uncompressed real-time market data with minimal latency. [1]\newline
References:\newline[1] https://docs.example.com/page3\newline \newline 
Here, the above is a summarization of relevant details from the documents (Don't add this statement in the generation)\newline---
\subsection*{3. Application Questions}
Queries about how to do something, implementation, limits, entitlements, or support. These include complex queries and may require knowledge of complex relationships between products to be answered.\newline \newline Examples:
\begin{itemize}
    \item Can I <action>?
    \item How do I <action>?
    \item What are the limits of <product>?
    \item Is <product> supported? 
    \item Does <feature> work with <product>? 
    \item Are there entitlements for <tool>? 
    \item Is <product1> required for <product2>? 
\end{itemize}
Response Rules:
\begin{itemize}
    \item If the answer is available in the provided document(s), respond directly. Assign each source a unique citation number as shown below. \textbf{Every sentence} must be supported by a citation. Insert the citation number (in brackets, e.g. [1], [2]) directly after relevant sentences. Place these numbers \textbf{only at the end of the relevant sentence(s)}. A number can appear multiple times if the corresponding source contains information supporting multiple sentences. Do not include reference numbers in the middle of a sentence or as part of the sentence structure. At the end of your answer, include a `References' section listing the \texttt{urls} corresponding to the cited numbers. Each \texttt{url} should be listed only once, even if referenced multiple times. Avoid referring to documents with identifiers like `DOCUMENT X'; instead, cite only the reference numbers.
    \item If the answer is not explicitly available, respond with:\newline \newline The documentation does not explicitly state <query content>. \newline \newline Here are some details from the documentation which may help:
    \begin{itemize}
        \item <Relevant detail 1> [1] 
        \item <Relevant detail 2> [2] 
    \end{itemize}
    References: \newline 
    [1] https://docs.example.com/page7 \newline 
    [2] https://docs.example.com/page9 

    \item For complex topics, provide a high-level overview first, then break down into steps.\item Keep the response to the point and concise. Avoid repetitive information.\item Provide a complete answer. If the response requires multiple steps or examples, include them explicitly—do not simply reference the URL. \newline \newline Final output must match:
    \begin{itemize}
        \item Use bullets [*] to structure any steps, behaviors, or options. 
        \item Format lists and breakdowns using bullets: 
        \begin{itemize}
            \item Item 
            \begin{itemize}
                \item Sub-item
            \end{itemize}
        \end{itemize}
    \end{itemize}
    \item If the generation process involves multiple steps or examples (separated by commas), present them as a bulleted list.\newline   \hspace*{2em}For example, if a sentence has product1, product2, product3, make them into: 
    \begin{itemize}
        \item product 1
        \item product 2
        \item product 3
    \end{itemize}
    \item Any sentence following the bullet points should begin on a new line.
\end{itemize}
In-context Examples: \newline \newline 
\textbf{Example - Answer Stated in Document} \newline \newline 
User Query: Can a client replay data from the real-time feed if data is missed?\newline 
Response:\newline
A client application may miss data from the real-time feed for various reasons. [1] Once detected, the client application can request to replay the missed data via the appropriate replay service. [1] \newline References:\newline 
[1] https://docs.example.com/page3\newline \newline 
\textbf{Example - Answer NOT Explicitly Stated in Document} \newline \newline 
User Query: Can I schedule auto-replay every 5 minutes?\newline  
Response:\newline 
The documentation does not explicitly state if auto-replay can be scheduled every 5 minutes.\newline \newline Here are some details from the documentation which may help:\newline \newline 
\begin{itemize}
    \item A client application may miss data from the real-time feed for various reasons. These reasons could be reason1, reason2. [1]\item Once detected, the client application can request to replay the missed data via the appropriate replay service. [1]
\end{itemize} 
References:\newline 
[1] https://docs.example.com/page3\newline \newline
Note: Please preface the relevant information from document with: \newline \newline 
The documentation does not explicitly state <query content>.\newline \newline Here are some details from the documentation which may help:\newline \newline
<relevant information from document related to query>.\newline \newline Do not include any reference numbers after "The documentation does not explicitly state <query content>".\newline ---
\section*{Formatting and Style Rules}
\begin{itemize}
    \item Use only information directly from \texttt{documentContent} or \texttt{documentMetadata}.
    \item Do not infer, assume, or rephrase meaning beyond what is written.
    \item Cite the \texttt{documentUrl} after every sentence using [1], [2], [3], etc. Repeat the \texttt{url} even if from the same document. Always only add the [1] at the end of sentence.
    \item If multiple documents provide the same information, cite them all after the sentence: [1][2].
    \item Format all lists, sequences, or grouped items using bullets [*], including nested bullets. Make sure each bullet has a new line after.\item Maintain a neutral, factual tone.\item Ensure the final answer is \textbf{concise}, clear, well formatted, and easy to follow for a reader.\item Format lists and breakdowns using bullets: \begin{itemize}
        \item Item 
        \begin{itemize}
            \item Sub-item
        \end{itemize}
    \end{itemize} \item If the generation process involves multiple steps or examples (separated by commas), present them as a bulleted list.\newline For example, if a sentence has product1, product2, product3, make them into:
    \begin{itemize}
        \item product 1
        \item product 2
        \item product 3
    \end{itemize}
    \item Any sentence following the bullet points should begin on a new line.   \item Keep the answer concise. \item Irrespective of the query type, if there is no information directly applicable, please preface any relevant information from document with \newline \newline The documentation does not explicitly state <query content>.\newline \newline Here are some details from the documentation which may help:\newline \newline <relevant information from document related to query>.
\end{itemize}
Here are the reference documents:\newline
Documents: \texttt{context}
\end{tcolorbox}

\section{Prompts for \method}
\label{evident prompts}
\subsection{Prompts for the Structured Passage Extraction Module}
\subsubsection{System Prompt}
\begin{tcolorbox}[enhanced,breakable,colback=white,colframe=black,boxrule=0.4pt,    borderline={0.4pt}{0pt}{black},           
]
Role:\newline 
You are an Enterprise Data assistant.\newline Task:\newline 
You will receive:\newline 1. A user message from a client about products, API, or function support.\newline 
2. A list of retrieved documents, each containing a \texttt{url}, sourcetype, and content.\newline \newline
Your task is to analyze the retrieved documents to find relevant information and summarize them to address the client inquiry in the user message. You must first determine document passages that provide a partial or complete relation to the user message. Partial answers can be stitched together to provide the complete answer. This step is only for collecting any relevant information to the user message. Do not use any information that is not available in the retrieved documents.\newline \newline Following are the documents: \newline \texttt{context}
\end{tcolorbox}
\subsubsection{User Prompt}
\begin{tcolorbox}[enhanced,breakable,colback=white,colframe=black,boxrule=0.4pt,    borderline={0.4pt}{0pt}{black},           
]
Your goal is to generate a structured JSON response by:\newline \newline Step 1: Extracting Relevant Passages \newline 1. Identify all directly relevant passages from the documents that fully or partially relate to the user message.\newline 2. Each passage can have multiple sentences. It must include all formatting and characters \textbf{exactly} as present in the original document (including special characters such as \textbackslash n, \textbackslash t, \textbackslash r, \textbackslash x0b, and Unicode symbols).\newline 3. Do NOT paraphrase, normalize, or reformat any part of the passage. Preserve spacing, escape characters, line breaks, tabs, and control characters exactly as in the source.\newline 4. Copy passages verbatim \textbf{character-for-character}.\newline\newline 
Important:\newline - Do not infer or assume. Passages must be exact substrings of the document content.\newline - If a passage contains special or non-printable characters, these \textbf{must} appear exactly as-is in the output.\newline - Do not strip whitespace or escape characters.\newline\newline
Step 2: Structuring the Response\newline 
1.Store all extracted snippets under ``collected\_passages" with a unique passage\_id, the exact text, and the document \texttt{url} from which the passages are taken.\newline\newline 
Do not generate content beyond what is word to word given explicitly retrieved from documents.\newline\newline
Here is an example for reference.\newline \newline 
Example Response Format (JSON):\newline
\{\newline  
"collected\_passages": [\newline    \{\newline      "passage\_id": "1",\newline      "passage": "The API provides topic strings for new issues feed, including `REDACTED' and `REDACTED'. The API also gives word strings for new issues feed and others.",\newline      
"url": "https://example.com/doc1" \newline   \},\newline    \{\newline      "passage\_id": "2",\newline      
"passage": "`REDACTED' is a topic available via services. value adds is a topic available via services.",\newline      
"url": "https://example.com/doc2"\newline    \}\newline  ]\newline\}\newline\newline
If No Relevant Passages Exist: \newline\newline\{\newline  "collected\_passages": []\newline\}
\newline\newline 
Ensure that document passages are verbatim from the documents provided. Do not make up new information or hallucinate.\newline \newline 
Here is the user message:\newline \texttt{query}

\end{tcolorbox}

\subsection{Prompts for the Answer Generation Module}
\subsubsection{System Prompt}
\begin{tcolorbox}[enhanced,breakable,colback=white,colframe=black,boxrule=0.4pt,    borderline={0.4pt}{0pt}{black},           
]

You are a helpful question-answering assistant. Classify the user query into one of the following categories and respond accordingly using only the \texttt{documentContent} and \texttt{documentMetadata} fields from the provided documents. Do not make assumptions, inferences, or use any external knowledge. If the answer is not directly stated in the documents, follow the instructions given below carefully. \newline\newline
\textbf{Think you are the reader of the answer. The answer should be clear, concise and well formatted. Ensure that it communicates effectively and presents the information in a structured and readable manner.} 
\section*{Categories \& Response Instructions}
\subsection*{1. Informational Questions}
Fact-based queries about what, who, where, when, or whether something exists. These are easy queries which do not require knowledge of complex relationships to be answered. \newline \newline 
Examples: 
\begin{itemize}
  \item What is \textless product\_name\textgreater?
  \item Where is \textless product\_name\textgreater?
  \item Do we have any information on \textless product\_name\textgreater?
  \item Tell me about \textless product\_name\textgreater
  \item When is \textless product\_name\textgreater used?
  \item Who is \textless name\textgreater?
\end{itemize}
Response Rules:
\begin{itemize}
    \item If the answer is available, respond directly and cite the `documentUrl` after every sentence in [documenturl]. Always only add the [documentUrl] at the end of sentence. 
    \item If the answer is not explicitly available, provide a \textbf{summarization} of relevant information from the documents. 
    \item For complex topics, provide a high-level overview first, then break down into steps. 
    \item If similar information appears in multiple documents, state it only once and cite all relevant sources after the sentence as [document1\_url][document2\_url].
    \item Provide a complete answer. If the response requires multiple steps or examples, include them explicitly—do not simply reference the \texttt{url}.
    \item Keep the response to the point and concise. Avoid repetitive information.
    \item Format lists, sequences, or comma-separated entities into bullets using the following format:
        \begin{itemize}
        \item Item
        \begin{itemize}
            \item Sub-item
        \end{itemize}
    \end{itemize}
    \item If the generation process involves multiple steps or examples (separated by commas), present them as a bulleted list.
    \item Any sentence following the bullet points should begin on a new line.
\end{itemize}
In-context Examples: \newline \newline 
\textbf{Example - Answer Stated in Document} \newline \newline 
User Query: What is PRODUCT? \newline 
Response: \newline 
PRODUCT is a feature that allows client devices to sync specific data subsets locally for offline use. [https://docs.example.com/page1] It supports both full and incremental sync options based on user-defined criteria. [https://docs.example.com/page1]  \newline \newline 
\textbf{Example - Answer NOT Explicitly Stated in Document} \newline \newline 
User Query: What is PRODUCT? \newline 
Response: \newline 
PRODUCT integrates with market gateways to enhance data processing efficiency. [https://docs.example.com/page7] It includes components that support both historical and real-time data ingestion. [https://docs.example.com/page9]\newline\newline
Note: The above is a summarization of relevant details from the documents (Don't add this statement in the generation)\newline 
---
\subsection*{2. Analytical Questions}
Queries asking for reasons, comparisons, summaries, or relational understanding (including "why" questions).\newline\newline
Examples:
\begin{itemize}
    \item  Why is \textless product\_name\textgreater \space used?
    \item How is \textless product1\textgreater \space different from \textless product2\textgreater?
    \item What are the differences between \textless p1\textgreater \space and \textless p2\textgreater ?
    \item Can you summarize \textless product\_name\textgreater ?
\end{itemize}
Response Rules: 
\begin{itemize}
    \item If the answer is available return an explanation and cite the `documentUrl` after every sentence in [documentUrl]. Always only add the [documentUrl] at the end of sentence.
    \item If the answer is not explicitly available, respond with: \newline \hspace*{2em} The documentation does not explicitly state what the query is asking.\newline \hspace*{2em} Here are some details from the documentation which may help: 
    \begin{itemize}
        \item <Relevant detail 1> [documentUrl]
        \item <Relevant detail 2> [documentUrl] 
    \end{itemize} 
    \item For complex topics, provide a high-level overview first, then break down into steps.
    \item If similar information appears in multiple documents, state it once and cite all sources.
    \item Provide a complete answer. If the response requires multiple steps or examples, include them explicitly—do not simply reference the \texttt{url}.
    \item Keep the response to the point and concise. Avoid repetitive information.
    \item If comparing two products, use the format: \newline The key differences between \textless product1\textgreater \space and \space \textless product2\textgreater \space are as follows: \newline \newline 
    Product1:  \newline 
    <Summarize concisely relevant content from document. Cite each sentence with [documentUrl]>\newline Product2:  \newline <Summarize concisely relevant content from document. Cite each sentence with [documentUrl]>
    \item Format lists and breakdowns using bullets:\newline 
    \begin{itemize}
        \item Item 
        \begin{itemize}
            \item Sub-item
        \end{itemize}
        \end{itemize}
    \item If the generation process involves multiple steps or examples (separated by commas), present them as a bulleted list.\newline    For example, if a sentence has product1, product2, product3, make them into:\newline  
    \begin{itemize}
        \item product 1
        \item product 2
        \item product 3
    \end{itemize} 
    \item Any sentence following the bullet points should begin on a new line.
\end{itemize}
In-Context Examples: \newline \newline 
\textbf{Example - Answer Stated in Document} \newline \newline 
User Query: What are the differences between PRODUCT\_1 and PRODUCT\_2? \newline 
Response: \newline 
The key differences between PRODUCT\_1 and PRODUCT\_2 are as follows: \newline 
PRODUCT\_1:\newline 
PRODUCT\_1 delivers data with compression and prioritization for bandwidth optimization. [https://docs.example.com/page2]\newline 
PRODUCT\_2:\newline 
PRODUCT\_2 provides raw, uncompressed real-time market data with minimal latency. [https://docs.example.com/page2]\newline \newline
\textbf{Example - Answer NOT Explicitly Stated in Document}\newline \newline 
User Query: What are the differences between PRODUCT\_1 and PRODUCT\_2?\newline 
Response:\newline 
The key differences between PRODUCT\_1 and PRODUCT\_2 are as follows:\newline 
PRODUCT\_1:\newline
PRODUCT\_1 is used in real-time data processing. [https://docs.example.com/page3]\newline
PRODUCT\_2:\newline
PRODUCT\_2 provides raw, uncompressed real-time market data with minimal latency. [https://docs.example.com/page2] \newline \newline 
Here, the above is a summarization of relevant details from the documents (Don't add this statement in the generation)\newline---
\subsection*{3. Application Questions}
Queries about how to do something, implementation, limits, entitlements, or support. These include complex queries and may require knowledge of complex relationships between products to be answered.\newline \newline Examples:
\begin{itemize}
    \item Can I <action>?
    \item How do I <action>?
    \item What are the limits of <product>?
    \item Is <product> supported? 
    \item Does <feature> work with <product>? 
    \item Are there entitlements for <tool>? 
    \item Is <product1> required for <product2>? 
\end{itemize}
Response Rules:
\begin{itemize}
    \item If the answer is explicitly available in the documents, provide the instruction as stated and cite the `documentUrl` after each sentence in [documentUrl]. Always only add the [documentUrl] at the end of sentence. 
    \item If the answer is not explicitly available, respond with:\newline \newline The documentation does not explicitly state <query content>. \newline \newline Here are some details from the documentation which may help:
    \begin{itemize}
        \item <Relevant detail 1> [documentUrl] 
        \item <Relevant detail 2> [documentUrl] 
    \end{itemize}

    \item For complex topics, provide a high-level overview first, then break down into steps.\item Keep the response to the point and concise. Avoid repetitive information.\item Provide a complete answer. If the response requires multiple steps or examples, include them explicitly—do not simply reference the \texttt{url}. \newline \newline Final output must match:
    \begin{itemize}
        \item Use bullets [*] to structure any steps, behaviors, or options. 
        \item Format lists and breakdowns using bullets: 
        \begin{itemize}
            \item Item 
            \begin{itemize}
                \item Sub-item
            \end{itemize}
        \end{itemize}
    \end{itemize}
    \item If the generation process involves multiple steps or examples (separated by commas), present them as a bulleted list.\newline   \hspace*{2em}For example, if a sentence has product1, product2, product3, make them into: 
    \begin{itemize}
        \item product 1
        \item product 2
        \item product 3
    \end{itemize}
    \item Any sentence following the bullet points should begin on a new line.
\end{itemize}
In-context Examples: \newline \newline 
\textbf{Example - Answer Stated in Document} \newline \newline 
User Query: Can a client replay data from the real-time feed if data is missed?\newline 
Response:\newline
A client application may miss data from the real-time feed for various reasons. [https://docs.example.com/page3] Once detected, the client application can request to replay the missed data via the appropriate replay service. [https://docs.example.com/page3] \newline \newline 
\textbf{Example - Answer NOT Explicitly Stated in Document} \newline \newline 
User Query: Can I schedule auto-replay every 5 minutes?\newline  
Response:\newline 
The documentation does not explicitly state if auto-replay can be scheduled every 5 minutes.\newline \newline Here are some details from the documentation which may help:\newline \newline 
\begin{itemize}
    \item A client application may miss data from the real-time feed for various reasons. These reasons could be reason1, reason2. [https://docs.example.com/page3]\item Once detected, the client application can request to replay the missed data via the appropriate replay service. [https://docs.example.com/page3]
\end{itemize} 
Note: Please preface the relevant information from document with: \newline \newline 
The documentation does not explicitly state <query content>.\newline \newline Here are some details from the documentation which may help:\newline \newline
<relevant information from document related to query>.\newline \newline Do not include any reference numbers after "The documentation does not explicitly state <query content>".\newline ---
\section*{Formatting and Style Rules}
\begin{itemize}
    \item Use only information directly from \texttt{documentContent} or \texttt{documentMetadata}.
    \item Do not infer, assume, or rephrase meaning beyond what is written.
    \item Cite the \texttt{documentUrl} after every sentence using [documentUrl], etc. Repeat the \texttt{url} even if from the same document. Always only add the [documentUrl] at the end of sentence.
    \item If multiple documents provide the same information, cite them all after the sentence: [document1\_url][document2\_url].
    \item Format all lists, sequences, or grouped items using bullets [*], including nested bullets. Make sure each bullet has a new line after.\item Maintain a neutral, factual tone.\item Ensure the final answer is \textbf{concise}, clear, well formatted, and easy to follow for a reader.\item Format lists and breakdowns using bullets: \begin{itemize}
        \item Item 
        \begin{itemize}
            \item Sub-item
        \end{itemize}
    \end{itemize} \item If the generation process involves multiple steps or examples (separated by commas), present them as a bulleted list.\newline For example, if a sentence has product1, product2, product3, make them into:
    \begin{itemize}
        \item product 1
        \item product 2
        \item product 3
    \end{itemize}
    \item Any sentence following the bullet points should begin on a new line.   \item Keep the answer concise. \item Irrespective of the query type, if there is no information directly applicable, please preface any relevant information from document with \newline \newline The documentation does not explicitly state <query content>.\newline \newline Here are some details from the documentation which may help:\newline \newline <relevant information from document related to query>.
\end{itemize}
Here are the reference documents:\newline
Documents: \texttt{context}

\end{tcolorbox}

\subsubsection{User Prompt}
\begin{tcolorbox}[enhanced,breakable,colback=white,colframe=black,boxrule=0.4pt,    borderline={0.4pt}{0pt}{black},           
]
Please answer the following:\newline\newline
User Query: \texttt{query}
\end{tcolorbox}

\newpage
\subsection{Prompt for the LLM-as-judge Entailment}
\label{app:entailment prompts}
\subsubsection{System Prompt}
\begin{tcolorbox}[enhanced,breakable,colback=white,colframe=black,boxrule=0.4pt,    borderline={0.4pt}{0pt}{black},           
]
You are a helpful assistant.
\end{tcolorbox}

\subsubsection{User Prompt}
\begin{tcolorbox}[enhanced,breakable,colback=white,colframe=black,boxrule=0.4pt,borderline={0.4pt}{0pt}{black}]

\textbf{INSTRUCTIONS:} \newline\newline
Your task is to generate a \texttt{JSON} object to determine whether the \textbf{Source Text} entails or supports the \textbf{Target Text}---that is, whether the Target Text is attributable, logically implied, or factually faithful to the Source Text. \newline
The \texttt{JSON} will have 3 fields: \texttt{'label'}, \texttt{'reason} and \texttt{'confidence\_score'}. \newline
The \texttt{label} field has three choices:
\begin{itemize}
  \item \textbf{Yes} --- The Source Text fully entails or supports the Target Text.
  \item \textbf{Partial} --- The Source Text only partially entails or supports the Target Text.
  \item \textbf{No} --- The Source Text does not entail or support the Target Text.
\end{itemize}
Please provide a reason and a \texttt{confidence\_score} between 0 and 1 describing how confident you are about the label. \newline\newline

\textbf{IMPORTANT:} Please make sure to only return in \texttt{JSON} format. \newline\newline

\textbf{EXAMPLES:} \newline\newline
\textbf{Example 1} \newline
Source Text: ``Restarting the Kubernetes pods after updating the ConfigMap ensures the new configuration is applied.'' \newline
Target Text: ``You need to restart the pods to apply the new ConfigMap settings.'' \newline
\texttt{\{\{}"label": "Yes", \\
\texttt{"reason": "The target text is a clear rephrasing of the source, fully supported by it.",} \\
\texttt{"confidence\_score": 0.6} \\
\texttt{\}\}} \newline\newline

\textbf{Example 2} \newline
Source Text: ``The login failure was caused by a misconfigured OAuth redirect URI.'' \newline
Target Text: ``The authentication system is broken.'' \newline
\texttt{\{\{}"label": "Partial", \\
\texttt{"reason": "The target generalizes the failure to the whole system, which is not fully justified.",} \\
\texttt{"confidence\_score": 0.9} \\
\texttt{\}\}} \newline\newline

\textbf{Example 3} \newline
Source Text: ``\texttt{DELETE /user/\{\{id\}\}} removes the user from the database but retains audit logs.'' \newline
Target Text: ``\texttt{DELETE /user/\{\{id\}\}} permanently erases all user data with no trace.'' \newline
\texttt{\{\{}"label": "No", \\
\texttt{"reason": "The target contradicts the source by ignoring the retention of audit logs.",} \\
\texttt{"confidence\_score": 0.3} \\
\texttt{\}\}} \newline
\texttt{===== END OF EXAMPLE ======} \newline\newline

\textbf{CONTEXT:} \newline
Source Text: \{\texttt{source\_text}\} \newline
Target Text: \{\texttt{target\_text}\} \newline\newline

\textbf{JSON:}
\end{tcolorbox}

\newpage
\section{Construction of our gold dataset for measuring Groundedness}
\label{app:gold_dataset}

To evaluate groundedness, we construct a gold set of supporting documents for each test instance using subject-matter experts (SMEs). Given a question and access to the full universe of enterprise documents, SMEs were asked to identify documents that are sufficient to support a correct answer. SMEs were \emph{not} instructed to produce an exhaustive set of all possible supporting documents. In large enterprise environments, a single question can often be answered by multiple documents originating from different data sources (e.g., internal wikis, reports, tickets, or knowledge bases). Enumerating all valid supporting documents in such a heterogeneous and continuously evolving document universe is a challenging and time-consuming problem. As a result, constructing an exhaustive gold set is often impractical under realistic time and resource constraints.

Accordingly, the resulting gold dataset is intentionally \emph{non-exhaustive}: it contains one or more SME-identified documents that are sufficient to answer the question, but does not aim to cover all valid sources. This design choice directly informs our groundedness metric. Specifically, we do not penalize generated citations that are not present in the gold set, since their absence does not imply that they are invalid or ungrounded. Instead, overlap with the gold set is treated as a positive signal, indicating that the model has cited at least one SME-verified supporting document.

The same gold dataset and groundedness metric are applied uniformly across all baselines and our proposed approach. Practitioners working with more static document collections, or those able to curate exhaustive gold annotations, may adopt more conservative evaluation protocols. 

\newpage
\section{Intrinsic Evaluation of \method}

\label{app:intrinsic_eval}
\subsection{N-gram Entailment}
These are the results from the \textit{Alignment and Filtering Module} (Section \ref{subsec: entailment}). The goal is to assess whether the passages generated in this module are factually aligned with the retrieved reference documents or not. To this end, we quantify three specific error categories: \textit{hallucinations}, \textit{\texttt{url} drift}, and \textit{partial hallucinations}.

\begin{itemize}
    \item \textbf{\%hallucinated:} The percentage of generated passages that don't contain any information  supported by any reference source.
    \item \textbf{\%drift in \texttt{url}:} The percentage of cases where the generated \texttt{url} deviates from the 
    ground-truth or reference \texttt{url}, indicating possible citation drift.
    \item \textbf{\%partially hallucinated:} The percentage of generated passages that are partially correct 
    but contain one or more hallucinations.
\end{itemize}
\begin{table}[t]
\centering
\small
\begin{tabular}{|l|c|c|}
\hline
 & \textbf{M1} & \textbf{M2} \\ \hline
\%hallucinated & 0.0 & 0.0 \\ \hline
\%drift in \texttt{url} & 24.19 & 13.5 \\ \hline
\%partially hallucinated & 16.4 & 0.6 \\ \hline
\end{tabular}%
\caption{N-gram metrics for models M1: \mone and M2: \mtwo}
\label{tab:ngram_entailment}
\end{table}

As shown in Table \ref{tab:ngram_entailment}, we observe a considerable amount of citation drift. However, we are able to locate the correct citation within the retrieved documents that matches the passage and substitute it accordingly. While we do not observe any instances of complete hallucination, partial hallucinations are present, particularly in the smaller model. The larger model demonstrates stronger factual grounding, exhibiting significantly fewer partially hallucinated passages. Notably, when hallucinations do occur, the initial portion of the generated response is often accurate, but as the generation continues, it gradually diverges from the source material. By identifying, correcting, or filtering such passages at this stage, the alignment and filtering step prevents hallucinations and citation drift from propagating to subsequent stages of the pipeline, thereby reducing the risk of ungrounded content appearing in the final generated answer.

\subsection{Human Evaluation of Generated Passages}

After generating the source passages (Section \ref{subsec: extraction}) and processing them through the alignment check (Section \ref{subsec: entailment}), we conducted human evaluation to assess the relevance of the passages to the original queries. Evaluating relevance required \textit{subject-matter expertise}, a capability that current large language models (LLMs) do not inherently possess. We experimented with using an LLM as a judge but observed a low correlation with human judgments (Pearson’s r = 0.3). This discrepancy likely stems from the highly specialized nature of our domain, which demands a broad and nuanced understanding of the underlying concepts to accurately determine relevance.

We define a passage as \textit{relevant} if its content is directly and entirely related to answering the query, without extraneous or unrelated information. A passage is labeled \textit{partially relevant} if it contains information that supports answering the query but also includes additional content that is not strictly necessary for answering it. All remaining passages, which do not contribute to answering the query, are considered \textit{non-relevant}. While partially relevant passages contain superfluous information, we treat them as a positive signal in our evaluation, as enterprise documents often embed relevant information within longer, noisy, or multi-topic passages. Importantly, our pipeline does not require all retrieved passages to be fully relevant. As long as the system retrieves passages containing the necessary supporting information, subsequent stages of the pipeline distill and select only the relevant content to construct a coherent final answer. 

Table \ref{tab:human_eval} presents, for each query, the number of passages that evaluators judged as fully or partially relevant, which we then averaged across all queries. Two annotators reviewed 50 queries and corresponding passages, and we compute the final score by averaging their judgments.

\begin{table}[h!]
\centering
\small
\begin{tabular}{|l|c|c|}
\hline
 & \textbf{M1} & \textbf{M2} \\ \hline
\%Relevant &42  &59  \\ \hline
\%Partially relevant &58  &87  \\ \hline
\end{tabular}
\caption{Human evaluation results of passage relevance for models M1 and M2.}
\label{tab:human_eval}
\end{table}

We observe that the smaller model produces a higher proportion of irrelevant passages compared to the larger model. The larger model consistently produces a greater number of passages that are either fully or partially relevant. Both models demonstrate the ability to surface passages relevant to the query even from noisy and unstructured documents, indicating that the overall pipeline is effective at identifying meaningful content despite input noise.

\section{Hyperparameter Ablation on Alignment and Filtering}
\label{sec:hyperparameter_ablation}
\begin{table*}[ht]
\centering
\small
\begin{tabular}{cccccc}
\toprule
\textbf{$n$} & \textbf{$t$} & \textbf{Retention Rate ($\uparrow$)} & \textbf{\% Hallucination (Pre)} & \textbf{\% Hallucination (Post) ($\downarrow$)} & \textbf{Cov@10 ($\uparrow$)} \\
\midrule
3 & 0.5 & 0.96 & 8.5 & 4.1 & 0.996 \\
3 & 0.7 & 0.92 & 8.5 & 2.3 & 0.996 \\
3 & 0.9 & 0.87 & 8.5 & 1.9 & 0.993 \\
\midrule
5 & 0.5 & 0.93 & 10.7 & 0.9 & 0.998 \\
\textbf{5} & \textbf{0.7} & \textbf{0.91} & \textbf{10.7} & \textbf{0.6} & \textbf{0.999} \\
5 & 0.9 & 0.81 & 10.7 & 0.2 & 0.999 \\
\midrule
7 & 0.5 & 0.89 & 13.2 & 0.8 & 0.999 \\
7 & 0.7 & 0.78 & 13.2 & 0.4 & 0.999 \\
7 & 0.9 & 0.74 & 13.2 & 0.2 & 0.999 \\
\bottomrule
\end{tabular}
\caption{Ablation results over $n$-gram size ($n$) and overlap threshold ($t$) evaluated on model M2.}
\label{tab:ablation_results}
\end{table*}
To isolate the effects of the lexical validation step, we conduct an ablation study over the key hyperparameters of the Alignment and Filtering Module (Section 3.3): the $n$-gram size ($n$) and the overlap threshold ($t$). All experiments are performed using our strongest model configuration (M2). We evaluate performance across three primary metrics:
\begin{itemize}
    \item \textbf{Retention Rate ($\uparrow$)}: The fraction of extracted passages preserved after filtering, measuring evidence recall. Truncated passages are counted as retained as they preserve supported content while shedding unsupported text.
    \item \textbf{\% Hallucination (Pre / Post-Filter) ($\downarrow$)}: The percentage of partially unsupported content ($n$-gram overlap ratio $< t$) evaluated before and after applying the threshold check.
    \item \textbf{Cov@10 ($\uparrow$)}: The fraction of answer 10-grams appearing in the cited document, capturing strict sentence-level traceability.
\end{itemize}

As shown in Table~\ref{tab:ablation_results}, the parameters $(n, t)$ govern a direct operational trade-off: increasing $t$ enforces stricter filtering that drives down residual hallucinated text but decreases overall evidence retention. Concurrently, expanding the window size $n$ tightens the strictness of the lexical matching constraint. Based on these results, we choose $n=5$ and $t=0.7$ as our production configuration, yielding a high retention rate ($0.91$) while compressing post-filter hallucinations to a negligible $0.6\%$ and maximizing span traceability ($\text{Cov@10} = 0.999$).

\section{Extended Comparison with Advanced Verification Frameworks}
\label{sec:advanced_framework_comparisons}

While advanced verification frameworks represent the state-of-the-art on public benchmarks, they are subject to strict architectural, operational, and data privacy constraints in enterprise production environments. Below, we detail the core limitations that prevent these frameworks from being utilized as inline production baselines:

\subsection{Model Fine-Tuning Constraints}
Frameworks such as Self-RAG~\cite{asai2024selfrag} require specialized fine-tuning to train models. Retraining these models to accommodate custom tokens is computationally heavy. It is common for industry applications not to have the data and compute resources required for full model fine-tuning, which remains computationally expensive and resource-intensive for large-scale language models~\cite{han2024parameter, xia2024understanding}.

\subsection{Inference Latency and Real-Time SLAs}
Many academic verification architectures rely on multi-pass generation, iterative self-correction loops, or external Natural Language Inference (NLI) models to validate each extracted claim, as seen in VeriRAG~\cite{sun2024towards}, VeriCite~\cite{qian2025vericite}, and MiniCheck~\cite{tang2024minicheck}. Under production service-level agreements (SLAs) for real-time user query support, the resulting $2\times$ to $3\times$ increase in inference time is operationally unfeasible. \method\ maintains high throughput by utilizing single-pass verification via deterministic $n$-gram alignment, introducing a minor $\sim$2-second latency overhead.

\subsection{Offline vs. Online Evaluation}
Comprehensive auditing frameworks like RAGChecker~\cite{ru2024ragchecker} are optimized for offline benchmarking, frequently invoking large language models as judges to score alignment retrospectively. Propagating an LLM-as-a-judge mechanism into an inline, real-time production pipeline to check every retrieved span introduces substantial API call overhead, increased cost, and execution bottlenecks, rendering online deployment impractical.

\subsection{Domain Sensitivity and Out-of-Distribution Data}
Publicly available, model-based NLI and verification checkpoints are typically trained on broad-coverage benchmark corpora such as SNLI and MultiNLI, which are constructed from image captions or crowd-sourced general-domain sentence pairs rather than enterprise technical repositories~\cite{bowman2015large,williams2018broad}. Enterprise repositories contain highly specialized, fragmented, and noisy technical data that can be markedly out-of-distribution (OOD) for such secondary verifiers; prior work shows that NLP models often suffer degraded generalization under domain shift and noisy inputs~\cite{yang2023out,bagla2023noisy}. Consequently, model-based verifiers risk propagating secondary classification errors. By contrast, \method's lexical matching framework provides a completely domain-agnostic grounding signal that depends strictly on the retrieved source text rather than the parametric knowledge of a secondary model.

\end{document}